\documentclass[review]{iise}

\usepackage{amsmath}
\usepackage{graphicx}
\usepackage{geometry}
\usepackage{booktabs}
\usepackage{multirow}
\usepackage{caption}
\usepackage{subcaption}
\usepackage{tikz}
\usetikzlibrary{positioning}
\usepackage{fancyhdr}

\usetikzlibrary{shapes, arrows, positioning, shadows, fit, backgrounds, calc, shapes.geometric}

\begin{document}

\pagestyle{fancy}
\fancyhf{}               
\fancyfoot[C]{\thepage}    % page number centered in footer
\renewcommand{\headrulewidth}{0pt}
\renewcommand{\footrulewidth}{0pt}
\pagenumbering{arabic}
\setcounter{page}{1}

\thispagestyle{fancy}

\vspace{0.5in}

\begin{center}
\textbf{\fontsize{16}{20}\selectfont
A Curriculum-Based Deep Reinforcement Learning Framework for the Electric Vehicle Routing Problem}
\end{center}

\vspace{0.10in}
\begin{center}
{\fontsize{11}{13}\selectfont
\textbf{Mertcan Daysalilar}$^{1}$, \textbf{Fuat Uyguroglu}$^{2}$, \textbf{Gabriel Nicolosi}$^{3}$, \textbf{Adam Meyers}$^{1}$\\[2pt]
{\fontsize{10}{12}\selectfont
$^{1}$Industrial and Systems Engineering, University of Miami, Coral Gables, FL, USA\\
$^{2}$Faculty of Engineering, Cyprus International University 99258 Nicosia, North Cyprus, via Mersin 10, Turkey\\[2pt]
$^{3}$Engineering Management and Systems Engineering,  Missouri University of Science and Technology, Rolla ,MO, USA\\[2pt]
\texttt{mxd4222@miami.edu}, \texttt{fuyguroglu@ciu.edu.tr}, \texttt{gabrielnicolosi@mst.edu},
\texttt{axm8336@miami.edu}
}
}
\end{center}
\vspace{0.10in}

\begin{center}
\textbf{\fontsize{12}{14}\selectfont Abstract}
\end{center}

\vspace{0.1in}

\noindent
\fontsize{10}{12}\selectfont
The electric vehicle routing problem with time windows (EVRPTW) is a complex optimization problem in sustainable logistics, where routing decisions must minimize total travel distance, fleet size, and battery usage while satisfying strict customer time constraints. Although deep reinforcement learning (DRL) has shown great potential as an alternative to classical heuristics and exact solvers, existing DRL models often struggle to maintain training stability—failing to converge or generalize when constraints are dense. In this study, we propose a curriculum-based deep reinforcement learning (CB-DRL) framework designed to resolve this instability. The framework utilizes a structured three-phase curriculum that gradually increases problem complexity: the agent first learns distance and fleet optimization (Phase A), then battery management (Phase B), and finally the full EVRPTW (Phase C). To ensure stable learning across phases, the framework employs a modified proximal policy optimization algorithm with phase-specific hyperparameters, value and advantage clipping, and adaptive learning-rate scheduling. The policy network is built upon a heterogeneous graph attention encoder enhanced by global-local attention and feature-wise linear modulation. This specialized architecture explicitly captures the distinct properties of depots, customers, and charging stations. Trained exclusively on small instances with $N=10$ customers, the model demonstrates robust generalization to unseen instances ranging from $N=5$ to $N=100$, significantly outperforming standard baselines on medium-scale problems. Experimental results confirm that this curriculum-guided approach achieves high feasibility rates and competitive solution quality on out-of-distribution instances where standard DRL baselines fail, effectively bridging the gap between neural speed and operational reliability.

\vspace{12pt}

\noindent 
\textbf{\fontsize{12}{14}\selectfont Keywords}\\
Electric vehicle routing problem with time windows,  curriculum learning, combinatorial optimization, proximal policy optimization, constraint decomposition

% ==========================================
% SECTIONS
% ==========================================

\section{Introduction}
The electrification of logistics fleets is a critical step toward sustainable industrial operations and supply chains, yet it introduces significant operational challenges. Unlike internal combustion engine vehicles, electric vehicles (EVs) are constrained by significantly shorter driving ranges, nonlinear charging times, and the scarcity of recharging infrastructure \cite{lin2021, schneider2014}. When coupled with strict customer time windows \cite{daysalilar2023}, the electric vehicle routing problem with time windows (EVRPTW) becomes an NP-hard optimization problem. The primary objective in industrial applications is not only to minimize distance but to optimize the total operational cost, which heavily weights the minimization of fleet size due to the high capital expenditure of electric vehicles \cite{toth2014}.

Traditional solvers, such as branch-and-price or adaptive large neighborhood search, provide high-quality solutions but suffer from computational bottlenecks that limit their utility in real-time, dynamic dispatching scenarios \cite{schneider2014, solomon1987, toth2014}. Recently, deep reinforcement learning (DRL) has emerged as a promising alternative, capable of generating solutions in seconds once the model is trained. Foundational works by  Kool et al. \cite{kool2019}  and Nazari et al. \cite{nazari2018} demonstrated that attention-based neural networks can effectively solve the traveling salesman problem (TSP) and capacitated vehicle routing problem (CVRP). Building on this, recent research has begun applying DRL to electric vehicle constraints. For instance, Lin et al. \cite{lin2021} adapted the attention mechanism to capture EV-specific states, such as battery levels and charging station availability, while Wang et al. \cite{wang2025} explored attention-enhanced strategies to optimize energy consumption. Despite these advances, applying end-to-end RL to highly constrained variants remains an open challenge due to the sparsity of feasible solutions in the search space. 

A major limitation of standard ``end-to-end'' DRL models is their tendency to fail when trained on the EVRPTW from random initialization \cite{lin2021}. This instability stems from the highly constrained nature of the problem: random exploration rarely yields a feasible solution, resulting in a sparse reward signal where the agent receives almost no positive feedback to guide learning. Instead, the agent encounters frequent negative feedback due to constraint violations (e.g., a depleted battery or missed deadline), leading to unstable gradient updates and degenerate policies that avoid hard-to-serve customers to minimize violation penalties \cite{lin2021}. 

To overcome such convergence issues, curriculum learning  \cite{bengio2009} has been proposed to gradually increase training difficulty. In the context of combinatorial optimization, Lisicki et al. \cite{lisicki2020} and Zhang et al. \cite{zhang2022} demonstrated that training on smaller graph sizes before scaling to larger instances improves generalization for the TSP and CVRP. However, these approaches primarily address scaling problem size ($N$) rather than the complexity of operational constraints. In the context of EVRPTW, effective training requires disentangling the learning of routing topology from ensuring feasibility under complex constraints \cite{lin2021}; an agent must first learn to construct a set of feasible routes before it can learn to optimize delivery timing. 

To address this instability, we propose a robust, curriculum-based deep reinforcement learning (CB-DRL) framework. Our primary contribution is a three-phase constraint curriculum that stabilizes training by decomposing the problem complexity. Specifically, the agent sequentially learns routing topology (phase A), energy management (phase B), and scheduling (phase C). We investigate whether this structured progression allows a neural policy to achieve near-optimal performance and zero-shot generalization on benchmark instances, effectively solving the feasibility collapse that plagues standard approaches.

\section{Problem Formulation}
We define the EVRPTW on a complete graph $G=(V, E)$, where $V = \{0\}\cup \mathcal{C}\cup \mathcal{S}$ consists of a depot node $0$, customer set $\mathcal{C}$ with cardinality $|\mathcal{C}|=N$, and charging station set $\mathcal{S}$ with $|\mathcal{S}|=M$. Each edge $(i,j) \in E$ has an associated distance $d_{ij}$ and travel time $t_{ij}$. Each customer $i$ has a demand $q_i$, service time $s_i$, and hard time window $[e_i, l_i]$.

The objective is to find a set of routes to serve all customers that minimizes the total operational cost ($J$), defined as the weighted sum of the total travel distance and the size of the vehicle fleet used:
\begin{equation}\label{eqn:J}
    J = \sum_{k \in K} \sum_{(i,j) \in \tau_k} d_{ij} + \lambda \cdot |K|
\end{equation}
where $K$ denotes the set of active vehicles used and $\tau_k$ denotes the specific trajectory (sequence of edges) traversed by vehicle $k \in K$. This objective function $J$ serves as the reward signal for the constructed solution set. Here, $\lambda$ reflects the capital cost of activating an additional EV, making fleet size reduction a primary operational objective. All customers must be visited exactly once in exactly one route, and each route starts and ends at the depot. The constraints include the following: (i) the accumulated demand of customers visited by any single vehicle must not exceed the maximum capacity $Q_{load}$, (ii) a vehicle's battery state-of-charge must remain non negative with full recharging allowed at stations, and (iii) service at customer $i$ must begin within its time window $[e_i, l_i]$ \cite{lin2021}.

\section{Methodology}
\begin{figure}[t!]
    \centering
    \resizebox{1.0\linewidth}{!}{
        \begin{tikzpicture}[
            node distance=1.5cm and 2cm,
            >=stealth,
            font=\small,
            % --- Professional Styles ---
            block/.style={rectangle, draw=black!60, fill=white, rounded corners=4pt, drop shadow, align=center, minimum height=1.2cm, inner sep=6pt},
            phase/.style={rectangle, draw=blue!50!black!60, fill=blue!5, rounded corners=2pt, align=center, font=\footnotesize, minimum height=1.0cm, minimum width=3.5cm},
            container/.style={draw=black!60, rectangle, dashed, line width=0.8pt, inner sep=0.5cm, rounded corners=6pt, fill=gray!2},
            % Arrow Styles
            flow/.style={->, line width=1.2pt, rounded corners=3pt, black!80},
            dashed_flow/.style={->, line width=1pt, dashed, rounded corners=3pt, blue!60!black},
            % Bigger arrow-label fonts
            math_label/.style={midway, font=\normalsize\bfseries},
            on_arrow_label/.style={midway, fill=white, inner sep=1.5pt, font=\footnotesize\bfseries, text=red!70!black}
        ]
            % =================================================================
            % RIGHT SIDE: PPO LOOP
            % =================================================================
            \node[block, fill=red!10, text width=5.0cm] (ppo) {
                \textbf{PPO Algorithm} \\
                Objective: $\nabla_\theta J(\theta)$ \\
                Phase-Adaptive Params
            };

            \node[block, fill=yellow!10, below=2cm of ppo, text width=4.5cm] (agent) {
                \textbf{Neural Agent $\pi_\theta$} \\
                (Hetero-Graph Encoder) \\
                Action $a_t \sim \pi_\theta(\cdot | s_t)$
            };

            \node[block, fill=green!10, left=2cm of agent, text width=3.5cm, minimum height=2cm] (env) {
                \textbf{EVRPTW Environment} \\
                Dynamic Constraints \\
                Active Set $\mathcal{C}_k$
            };

            \begin{pgfonlayer}{background}
                \node[
                    container,
                    fit=(ppo) (agent) (env),
                    inner xsep=1.0cm,
                    inner ysep=0.9cm,
                    yshift=-0.6cm,
                    label=above:\textbf{Policy Optimization Loop}
                ] (right_container) {};
            \end{pgfonlayer}

            % =================================================================
            % LEFT SIDE: THE CURRICULUM
            % =================================================================
            \node[block, fill=orange!10, text width=4cm, left=7.5cm of ppo] (curric) {
                \textbf{Curriculum Controller} \\
                Selects Active $\mathcal{C}_k$
            };

            \draw[flow, black!60]
                ($(curric.north)+(0,0.5)$)
                -- node[right, font=\normalsize\bfseries] {Epoch $k$}
                (curric.north);

            \node[phase, below=0.8cm of curric] (phaseA) {
                \textbf{Phase A: Topology} \\
                $\min \sum d_{ij}$ s.t. Capacity $Q$
            };
            \node[phase, below=0.8cm of phaseA] (phaseB) {
                \textbf{Phase B: + Energy} \\
                s.t. Battery $b_t \ge 0$
            };
            \node[phase, below=0.8cm of phaseB] (phaseC) {
                \textbf{Phase C: + Time} \\
                s.t. $e_i \le a_i \le l_i$
            };

            \begin{pgfonlayer}{background}
                \node[
                    container,
                    fit=(curric) (phaseC),
                    label=above:\textbf{Constraint Decomposition}
                ] (left_container) {};
            \end{pgfonlayer}

            % =================================================================
            % CONNECTIONS
            % =================================================================
            \draw[flow, blue!60!black] (curric) -- (phaseA);
            \draw[flow, blue!60!black] (phaseA) -- (phaseB);
            \draw[flow, blue!60!black] (phaseB) -- (phaseC);

            \draw[dashed_flow] (phaseA.east) to[out=0,in=160]
                node[on_arrow_label, sloped] {$k < 10$} (env.west);
            \draw[dashed_flow] (phaseB.east) to[out=0,in=180]
                node[on_arrow_label, sloped] {$10 \le k < 20$} (env.west);
            \draw[dashed_flow] (phaseC.east) to[out=0,in=200]
                node[on_arrow_label, sloped] {$k \ge 20$} (env.west);

            \draw[flow] (env.east) -- node[above, math_label] {$s_t, r_t$} (agent.west);
            \draw[flow] (agent.south) |- ++(0,-0.8) -|
                node[near start, below, math_label] {Action $a_t$} (env.south);
            \draw[flow] (agent.north) -- node[right, math_label] {Trajectory $\tau$} (ppo.south);
            \draw[flow, red!80!black, dashed]
                (ppo.east) to[out=0, in=0, looseness=2]
                node[right, math_label] {Update $\theta$} (agent.east);

        \end{tikzpicture}
    }
    \caption{Overview of the proposed curriculum-based deep reinforcement learning (CB-DRL) framework.}
    \label{fig:framework}
    \vspace{-0.5em}
\end{figure}
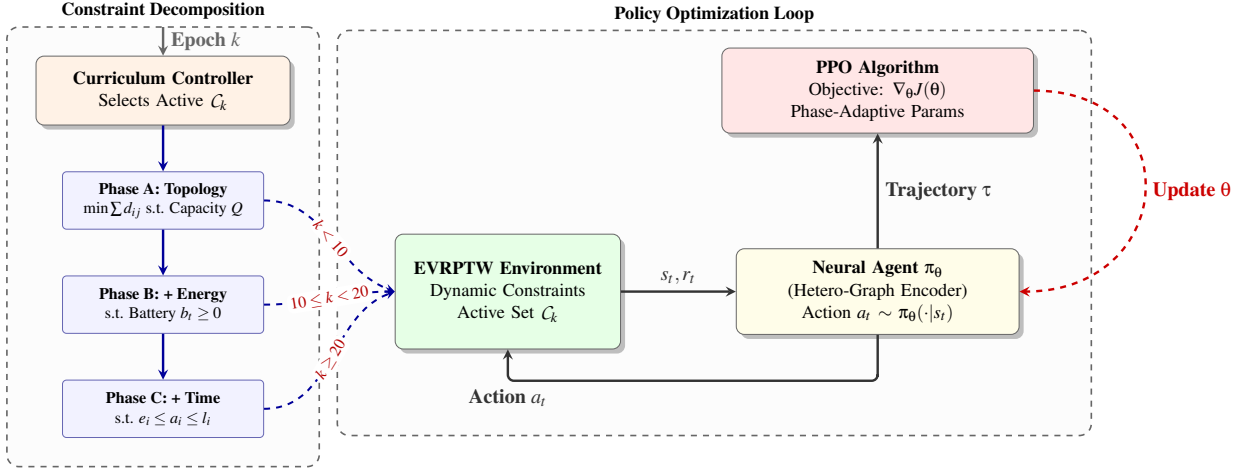

The proposed framework combines a heterogeneous encoder-decoder architecture with a curriculum-based training controller as shown in Fig. \ref{fig:framework}. The architecture consists of a constraint decomposition module and a policy optimization loop. The controller adjusts training complexity based on the current epoch $k$ by determining the active constraint set $\mathcal{C}_k$. This schedule divides training into three phases. Phase A ($k < 10$) enforces only topological and capacity limits, phase B ($10 \le k < 20$) injects energy constraints and battery limits, and phase C ($k \ge 20$) activates the full problem by enforcing time windows. These constraints define the transition dynamics and reward function within the environment. The neural agent $\pi_\theta$ interacts with this environment by observing state $s_t$ and generating action $a_t$. The environment returns reward $r_t$ based on operational costs and constraint violations. The resulting trajectories are used to update model parameters $\theta$.

\subsection{Heterogeneous Graph Attention Encoder}

Our neural policy network employs a heterogeneous graph attention encoder to effectively capture the distinct roles of different node types in the EVRPTW. Standard attention models treat all graph nodes identically, applying the same projection and attention mechanisms regardless of node type. However, in EVRPTW, nodes serve fundamentally different functions: charging stations provide battery replenishment (renewable resources), customers impose demand and time constraints (demand nodes), and the depot serves as the route origin and destination. To address this heterogeneity, and inspired by \cite{wang2025}, we employ separate projection parameters $W^Q_{cust}, W^Q_{station}$, and $W^Q_{depot}$. This allows the model to learn distinct relational dynamics—for example, that the distance between a customer and a station is critical for feasibility, whereas the distance between two stations is less relevant. The output embeddings are fed into a global-local attention edge encoder to fuse local neighborhood information with global routing context \cite{yan2025}, effectively aggregating features across different spatial scales.

\subsection{The Phased Constraint Curriculum}
The core novelty of this work is the curriculum controller, which modulates the environment's complexity to ensure stable convergence based on the principles of curriculum learning \cite{bengio2009}. We resolve ``gradient collapse'' by structuring training into three phases. Phase A (topology learning) enforces only capacity constraints ($Q_{load}$), while battery limits and time windows remain disabled. In this stage, the agent learns basic spatial routing and the ``return to depot'' logic, establishing a strong spatial foundation. Phase B (energy management) subsequently introduces battery-related ($Q_{bat}$) constraints and activates charging station nodes. During this phase, the agent learns to monitor its state of charge and detour to charging stations when necessary. Finally, phase C (full EVRPTW) enforces all constraints, including time windows $[e_i, l_i]$, requiring the agent to refine its routing and charging policies to strictly satisfy customer time windows. Each phase restricts the optimization problem to a reduced constraint set, yielding a simpler reward structure that mitigates conflicting optimization objectives (e.g., distance minimization versus time-window feasibility). By activating constraint-related penalties incrementally, the framework stabilizes policy-gradient updates and enables the agent to learn feasible routing behavior before optimizing more restrictive constraints.

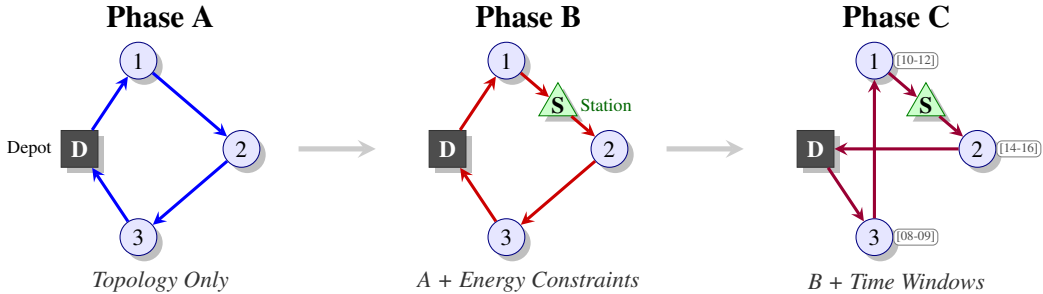
\begin{figure}[h]
    \centering
    \resizebox{0.85\linewidth}{!}{
        \begin{tikzpicture}[
            node distance=1.5cm,
            >=stealth,
            font=\small,
            % --- Node Styles ---
            depot/.style={rectangle, draw=black!80, fill=black!70, text=white, inner sep=2pt, minimum size=0.5cm, drop shadow},
            cust/.style={circle, draw=blue!50!black, fill=blue!10, inner sep=0pt, minimum size=0.5cm, drop shadow},
            station/.style={regular polygon, regular polygon sides=3, draw=green!50!black, fill=green!20, inner sep=-1.5pt, minimum size=0.6cm, drop shadow},
            % --- Label Styles ---
            tw_tag/.style={rectangle, draw=gray, fill=white, inner sep=1pt, rounded corners=2pt, font=\tiny, text=black!70},
            phase_title/.style={font=\bfseries\large, align=center, yshift=1.8cm},
            phase_desc/.style={font=\itshape\small, align=center, yshift=-1.8cm, text=black!80},
            % --- Route Styles ---
            route/.style={->, line width=1.2pt}, 
            arrow_trans/.style={->, line width=2pt, gray!40, shorten >=5pt, shorten <=5pt}
        ]

            % ==========================================
            % PHASE A: TOPOLOGY (Left)
            % ==========================================
            \begin{scope}[local bounding box=phaseA]
                \node[phase_title] at (1.1, 0) {Phase A};
                
                \node[depot, label=left:\scriptsize Depot] (D_a) at (0,0) {\textbf{D}};
                \node[cust] (C1_a) at (0.8, 1.2) {1};
                \node[cust] (C2_a) at (2.2, 0.0) {2};
                \node[cust] (C3_a) at (0.8, -1.2) {3};
                \draw[route, blue] (D_a) -- (C1_a);
                \draw[route, blue] (C1_a) -- (C2_a);
                \draw[route, blue] (C2_a) -- (C3_a);
                \draw[route, blue] (C3_a) -- (D_a);
                
                \node[phase_desc] at (1.1, 0) {Topology Only};
            \end{scope}

            \draw[arrow_trans] (2.8, 0) -- (4.2, 0);

            % ==========================================
            % PHASE B: ENERGY (Center)
            % ==========================================
            \begin{scope}[xshift=5.0cm, local bounding box=phaseB]
                \node[phase_title] at (1.1, 0) {Phase B};
            
                \node[depot] (D_b) at (0,0) {\textbf{D}};
                \node[cust] (C1_b) at (0.8, 1.2) {1};
                \node[cust] (C2_b) at (2.2, 0.0) {2};
                \node[cust] (C3_b) at (0.8, -1.2) {3};
                \node[station, label=right:\scriptsize \textcolor{green!40!black}{Station}] (S_b) at (1.5, 0.6) {\textbf{S}};
                
                % Split into individual segments
                \draw[route, red!80!black] (D_b) -- (C1_b);
                \draw[route, red!80!black] (C1_b) -- (S_b);
                \draw[route, red!80!black] (S_b) -- (C2_b);
                \draw[route, red!80!black] (C2_b) -- (C3_b);
                \draw[route, red!80!black] (C3_b) -- (D_b);
                
                \node[phase_desc] at (1.1, 0) {A + Energy Constraints};
            \end{scope}

            \draw[arrow_trans] (7.8, 0) -- (9.2, 0);

            % ==========================================
            % PHASE C: TIME WINDOWS (Right)
            % ==========================================
            \begin{scope}[xshift=10.0cm, local bounding box=phaseC]
                \node[phase_title] at (1.1, 0) {Phase C};
            
                \node[depot] (D_c) at (0,0) {\textbf{D}};
                \node[cust] (C1_c) at (0.8, 1.2) {1};
                \node[cust] (C2_c) at (2.2, 0.0) {2};
                \node[cust] (C3_c) at (0.8, -1.2) {3};
                \node[station] (S_c) at (1.5, 0.6) {\textbf{S}};
                
                \node[tw_tag, anchor=west] at (C1_c.east) {[10-12]};
                \node[tw_tag, anchor=west] at (C2_c.east) {[14-16]};
                \node[tw_tag, anchor=west] at (C3_c.east) {[08-09]};
                
                % Split into individual segments
                \draw[route, purple!80!black] (D_c) -- (C3_c);
                \draw[route, purple!80!black] (C3_c) -- (C1_c);
                \draw[route, purple!80!black] (C1_c) -- (S_c);
                \draw[route, purple!80!black] (S_c) -- (C2_c);
                \draw[route, purple!80!black] (C2_c) -- (D_c);
                
                \node[phase_desc] at (1.1, 0) {B + Time Windows};
            \end{scope}

        \end{tikzpicture}
    }
    \caption{The three-phase constraint curriculum.}
    \label{fig:curriculum}
    \vspace{-2.5em}
\end{figure}
\section{Experimental Design and Results}
\subsection{Experimental Setup}

We utilize a custom instance generator that produces feasible EVRPTW instances. Customer and charging-station coordinates are sampled uniformly from the unit square $[0,1]\times[0,1]$, following standard benchmarking practice \cite{kool2019, lin2021}, with travel distances and times proportional to Euclidean distance. The generator produces instances across nine spatiotemporal classes, denoted C, Cm, Ct, R, Rm, Rt, RC, RCm, and RCt. The spatial notation follows Solomon-style \cite{solomon1987} distributions: C (clustered customers), R (randomly distributed customers), and RC (mixed random--clustered). Temporal tightness is indicated with suffixes: wide (no suffix), medium (m), and tight (t) time windows. Customer demands are sampled from a bounded integer range, and vehicle capacity is fixed to ensure feasibility. Charging stations are placed independently and allow full recharging. 

To assess zero-shot generalization (i.e., evaluating performance on unseen instances without retraining), the model is trained \textit{only} on instances with $N=10$ customers and $M=3$ charging stations, and then evaluated without retraining on instance sizes $N\in\{5,10,20,30,40,50,100\}$. During the evaluation phase, to ensure robust performance, we employ the policy-oriented Monte Carlo decoding scheme \cite{kwon2020} with a fixed batch size of $N$ starting nodes.

We compare our curriculum-based deep reinforcement learning (CB-DRL) framework against three distinct baselines. First, we utilize the variable neighborhood search (VNS) heuristic proposed by Schneider et al. \cite{schneider2014} for the EVRPTW. Second, we include a standard end-to-end proximal policy optimization (PPO) model trained directly on the full EVRPTW without curriculum learning. Finally, we employ an exact optimization approach based on a mixed-integer linear programming (MILP) formulation of the EVRPTW \cite{daysalilar2023}. The MILP explicitly models routing, charging decisions, battery dynamics, and time-window constraints, and it is solved to optimality using the Gurobi Optimizer to provide ground-truth benchmarks and optimality gaps.

We evaluate solution quality and computational efficiency using four complementary metrics, including total cost ($J$), optimality gap ($\Delta\%$), feasibility success rate, and runtime. The optimality gap ($\Delta\%$) measures the relative deviation of the total operational cost from a baseline solution (exact, heuristic, or best found) and is computed as $(J_{\text{RL}} - J_{\text{base}})/J_{\text{base}} \times 100$, where $J$ (Eq. \ref{eqn:J}) includes total travel distance and fleet activation cost. In all experiments, the fleet activation penalty is fixed at $\lambda =100$. The feasibility success rate (\%) serves as the primary metric for assessing the framework's ability to consistently generate feasible solutions on unseen test instances under tight operational constraints. A strict time limit of 600 seconds was imposed on all solvers. Any method failing to converge within this limit—indicated by a hyphen (-) in the results—was considered incomplete.

\subsection{Solution Quality and Generalization}

Table \ref{tab:gap_analysis} details the generalization performance. We report the total travel distance ($D$)—corresponding to the first term of the objective function (Eq. \ref{eqn:J})—alongside the fleet size ($K$) and total operational cost ($J$). We observe that while both neural models perform comparably on small instances ($N < 20$), CB-DRL begins to outperform standard PPO on instances with $N \ge 20$, maintaining tighter optimality gaps as complexity increases.

\begin{table}[h]
\centering
\caption{Detailed performance breakdown including distance ($D$), fleet size ($K$), total cost ($J$), and optimality gap ($\Delta$)}
\label{tab:gap_analysis}
\scriptsize 
\resizebox{1.0\textwidth}{!}{
    \setlength{\tabcolsep}{3pt}
    \renewcommand{\arraystretch}{1.2}
    \begin{tabular}{l ccc cccc cccc cccc}
    \toprule
    \multirow{2}{*}{\textbf{Instance}} & 
    \multicolumn{3}{c}{\textbf{MILP (Exact)}} & 
    \multicolumn{4}{c}{\textbf{Heuristic (VNS)}} & 
    \multicolumn{4}{c}{\textbf{Standard PPO}} & 
    \multicolumn{4}{c}{\textbf{CB-DRL}} \\
    \cmidrule(r){2-4} \cmidrule(lr){5-8} \cmidrule(lr){9-12} \cmidrule(l){13-16} 
    & $D$ & $K$ & $J$ & $D$ & $K$ & $J$ & $\Delta\%$ 
    & $D$ & $K$ & $J$ & $\Delta\%$ 
    & $D$ & $K$ & $J$ & $\Delta\%$ \\
    \midrule
    \textbf{C5S2} & 2.19 & 1.13 & \textbf{115.2} & 2.20 & 1.15 & 117.2 & 1.7 & 2.04 & 1.20 & 122.0 & 5.9 & 2.03 & 1.20 & 122.0 & 5.9 \\
    \textbf{C10S3} & 3.22 & 2.04 & \textbf{207.2} & 3.26 & 2.10 & 213.3 & 2.9 & 3.30 & 2.14 & 217.3 & 4.9 & 3.28 & 2.14 & 217.3 & 4.9 \\
    \textbf{C20S3} & - & - & - & 5.19 & 3.69 & \textbf{374.2} & \textbf{0.0} & 7.29 & 3.75 & 382.3 & 2.2 & 5.68 & 3.72 & 377.7 & 0.9 \\
    \textbf{C30S4} & - & - & - & 5.90 & 3.98 & \textbf{403.9} & \textbf{0.0} & 10.1 & 4.12 & 422.1 & 4.5 & 6.91 & 4.13 & 419.9 & 4.0 \\
    \textbf{C40S5} & - & - & - & - & - & - & - & 13.3 & 4.41 & 454.3 & 5.0 & 8.59 & 4.28 & \textbf{436.6} & \textbf{0.0} \\
    \textbf{C50S6} & - & - & - & - & - & - & - & 16.4 & 4.82 & 498.4 & 6.0 & 10.4 & 4.61 & \textbf{471.4} & \textbf{0.0} \\
    \textbf{C100S12} & - & - & - & - & - & - & - & 33.1 & 8.01 & 834.1 & 18.5 & 18.2 & 7.22 & \textbf{740.2} & \textbf{0.0} \\
    \bottomrule
    \end{tabular}
}

\caption*{\scriptsize \textit{Note: Bold indicates the best solution ($J$) found across all methods. Baseline for $\Delta\%$ is MILP ($N \le 10$), heuristic ($20 \le N \le 30$), and best found solution ($N \ge 40$).}}
\end{table}

\begin{table}
\centering
\caption{Feasibility Success Rate and Runtime on Unseen Test Instances}
\label{tab:feasibility}
\small
\setlength{\tabcolsep}{2.5pt}
\begin{tabular}{l cc cc cc cc cc cc cc}
\toprule
 & \multicolumn{2}{c}{\textbf{N=5}} & \multicolumn{2}{c}{\textbf{N=10}} & \multicolumn{2}{c}{\textbf{N=20}} & \multicolumn{2}{c}{\textbf{N=30}} & \multicolumn{2}{c}{\textbf{N=40}} & \multicolumn{2}{c}{\textbf{N=50}} & \multicolumn{2}{c}{\textbf{N=100}} \\
\cmidrule(lr){2-3} \cmidrule(lr){4-5} \cmidrule(lr){6-7} \cmidrule(lr){8-9} \cmidrule(lr){10-11} \cmidrule(lr){12-13} \cmidrule(lr){14-15}
\textbf{Model} & \textbf{Succ.} & \textbf{Time} & \textbf{Succ.} & \textbf{Time} & \textbf{Succ.} & \textbf{Time} & \textbf{Succ.} & \textbf{Time} & \textbf{Succ.} & \textbf{Time} & \textbf{Succ.} & \textbf{Time} & \textbf{Succ.} & \textbf{Time} \\
\midrule
MILP (Exact)     & \textbf{100}\% & 0.21s & \textbf{100}\% & 64.4s & - & - & - & - & - & - & - & - & - & - \\
Heuristic (VNS)       & 97.6\% & 0.36s & 99.6\% & 0.86s & \textbf{98.2}\% & 3.75s & \textbf{98.4}\% & 11.52s & - & - & - & - & - & - \\
Standard PPO     & 94.8\% & 0.28s & 96.2\% & 1.12s & 92.1\% & 4.74s & 78.3\% & 11.54s & 66.2\% & 23.86s & 61.1\% & 35.91s & 64.4\% & 188.48s \\
\textbf{CB-DRL} & 94.5\% & 0.31s & 95.5\% & 1.25s & 92.3\% & 5.15s & 87.4\% & 11.26s & \textbf{74.7\%} & 20.64s & \textbf{68.9\%} & 32.10s & \textbf{64.7\%} & 142.19s \\
\bottomrule
\end{tabular}
{\captionsetup{skip=2pt}\caption*{\footnotesize \textit{Succ. = Success rate (\% feasible solutions). Time = Average inference time per instance.}}}
\vspace{-1.0em}
\end{table}

As problem complexity increases ($N \ge 40$), CB-DRL achieves higher optimality than the baselines, with the exact solver and heuristic timing out at $N\geq 20$ and $N\geq 40$ instance sizes, respectively. As detailed in Table \ref{tab:gap_analysis}, CB-DRL gives the best-known solutions (bold) for $N=40$, $50$, and $100$, significantly outperforming standard PPO, which suffers from increasingly large optimality gaps (e.g., an 18.5\% gap at $N=100$). This advantage is reinforced by the feasibility success rates in Table \ref{tab:feasibility}; on unseen large-scale instances ($N \ge 40$), CB-DRL consistently maintains a higher success rate than standard PPO (e.g., 74.7\% vs. 66.2\% at $N=40$). Finally, CB-DRL shows improved computational efficiency; at $N=100$, it is approximately 25\% faster than standard PPO (142.19s vs. 188.48s) because it generates concise, feasible routes rather than the excessively long, detoured paths produced by the baseline (e.g., Fig. 3(c-d)).

\begin{figure}[htbp]
    \centering
    % Top image (Metrics)
    \includegraphics[width=1.0\linewidth]{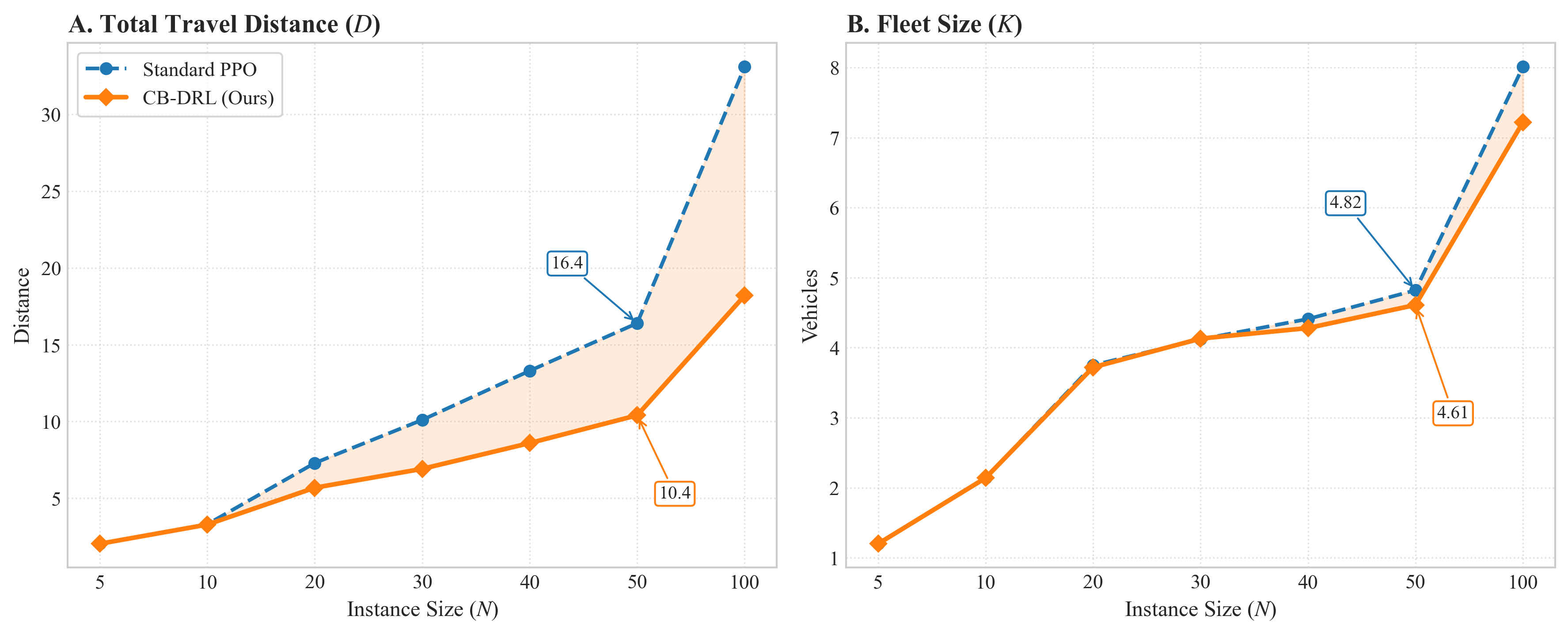}
    
    \vspace{0.1cm} 
    
    % Bottom image (Maps)
    \includegraphics[width=0.95\linewidth]{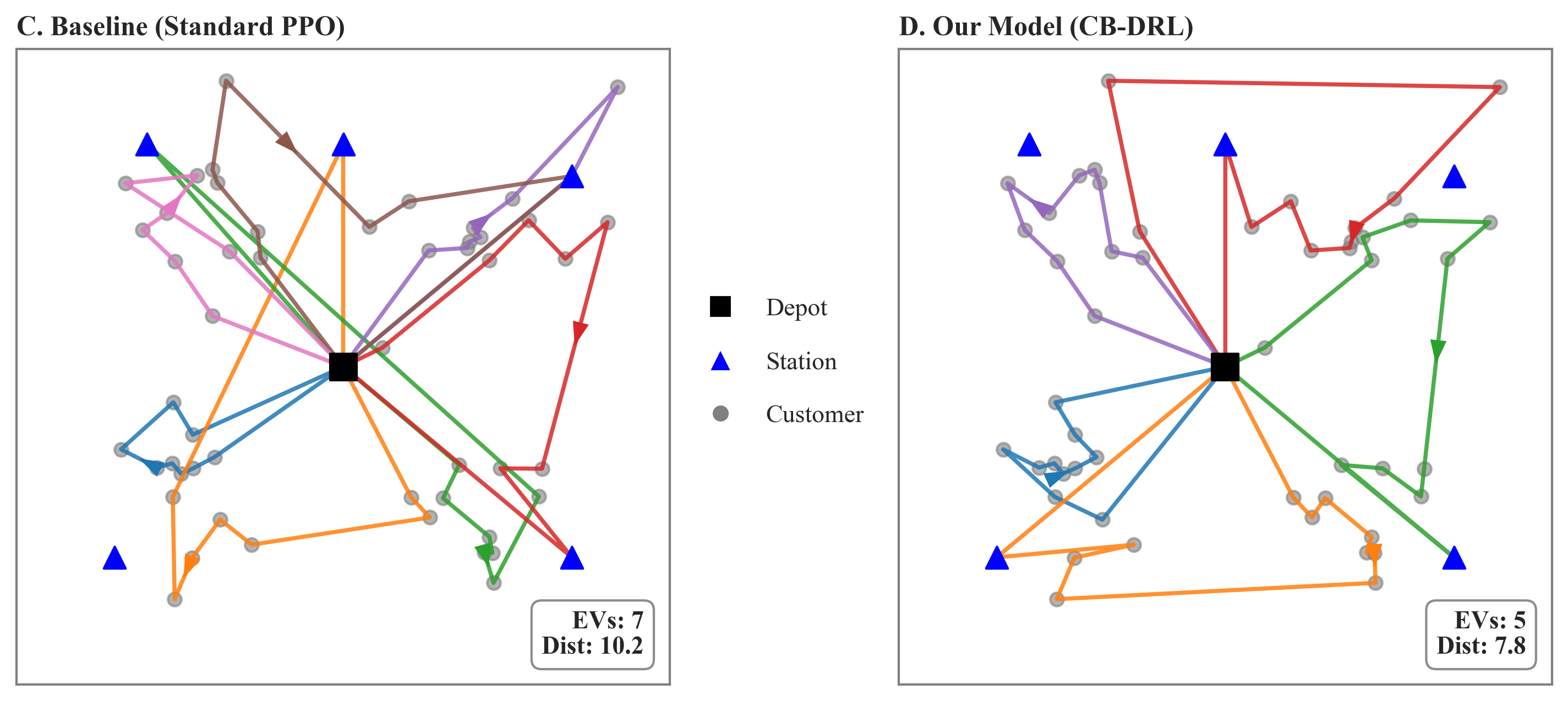}
    
    \vspace{-0.2cm} 
    
    \caption{Performance metrics (A, B) and routing behavior (C, D) comparing standard PPO and CB-DRL.}
    \label{fig:behavior}
    
    \vspace{-2.5cm} 
\end{figure}
\newpage
Figure \ref{fig:behavior} illustrates the behavioral mechanism driving these results. Quantitative metrics (top) confirm that the performance advantage of CB-DRL grows with problem scale. Qualitatively, the route maps (bottom)—visualizing a representative instance with $N=40$ customers and $M=5$ charging stations—reveal that the baseline model struggles with long-term planning, resulting in tangled, inefficient trajectories. In contrast, the curriculum-trained agent generates clean, well-separated routes, minimizing unnecessary detours and allowing it to serve the same demand with a smaller fleet.

\section{Conclusion}
This study addresses the limitations of standard deep reinforcement learning in solving highly constrained optimization problems. We proposed a curriculum-based DRL (CB-DRL) framework that structurally decomposes the electric vehicle routing problem with time windows into a learnable hierarchy of topology, energy, and time. Our experimental results highlight three primary contributions regarding scalability and performance.

First, the framework demonstrates superior scalability compared to traditional methods; while the exact solver and heuristic become intractable or exceed time limits for large problem sizes ($N \ge 40$), CB-DRL efficiently generates valid solutions up to $N=100$. Second, this scalability translates to superior solution quality on large instances ($N \ge 40$), where the model achieves higher optimality and feasibility success rates than the standard PPO baseline. Third, the proposed method improves computational efficiency, yielding faster inference times on instances of $N \ge 30$ by generating concise routes that minimize decoding steps. These findings suggest that for complex combinatorial tasks, introducing constraints sequentially allows neural solvers to learn robust foundational policies before tackling full operational complexity.

Future research will focus on extending the CB-DRL framework to stochastic environments, incorporating uncertain travel times and dynamic customer requests to better reflect real-world operational volatility. Additionally, we aim to investigate the integration of this neural policy into meta-heuristic search operators, enabling scalable solutions for massive-scale logistics networks ($N \ge 500$) without sacrificing computational efficiency.

% ==========================================
% REFERENCES
% ==========================================
\renewcommand\refname{\fontsize{12}{14}\selectfont References}

\end{document}